\documentclass[conference]{IEEEtran}
\IEEEoverridecommandlockouts
\usepackage{cite}
\usepackage{amsmath,amssymb,amsfonts}
\usepackage{algorithmic}
\usepackage{graphicx}
\usepackage{textcomp}
\usepackage{xcolor}
\usepackage{multirow}
\usepackage[hidelinks]{hyperref}
\usepackage{threeparttable}

\def\BibTeX{{\rm B\kern-.05em{\sc i\kern-.025em b}\kern-.08em
    T\kern-.1667em\lower.7ex\hbox{E}\kern-.125emX}}
\begin{document}

\title{Mental Multi-class Classification on Social Media: Benchmarking Transformer Architectures against LSTM Models}

\author{
    \IEEEauthorblockN{Khalid Hasan}
    \IEEEauthorblockA{
        \textit{Dept. of Computer Science} \\
        \textit{Missouri State University}\\
        Springfield, MO, USA \\
        kh597s@missouristate.edu
    }
    \and
    \IEEEauthorblockN{Jamil Saquer}
    \IEEEauthorblockA{
        \textit{Dept. of Computer Science} \\
        \textit{Missouri State University}\\
        Springfield, MO, USA \\
        jamilsaquer@missouristate.edu
    }
    \and
    \IEEEauthorblockN{Yifan Zhang}
    \IEEEauthorblockA{
        \textit{Dept. of Computer Science} \\
        \textit{Missouri State University}\\
        Springfield, MO, USA \\
        yifanzhang@missouristate.edu
    }
}

\maketitle

\begin{abstract}

Millions of people openly share mental health struggles on social media, providing rich data for early detection of conditions such as depression, bipolar disorder, etc. However, most prior Natural Language Processing (NLP) research has focused on single-disorder identification, leaving a gap in understanding the efficacy of advanced NLP techniques for distinguishing among multiple mental health conditions.  
In this work, we present a large-scale comparative study of state-of-the-art transformer versus Long Short-Term Memory (LSTM)-based models to classify mental health posts into exclusive categories of mental health conditions.  
We first curate a large dataset of Reddit posts spanning six mental health conditions and a control group, using rigorous filtering and statistical exploratory analysis to ensure annotation quality. We then evaluate five transformer architectures (BERT, RoBERTa, DistilBERT, ALBERT, and ELECTRA) against several LSTM variants (with or without attention, using contextual or static embeddings) under identical conditions. Experimental results show that transformer models consistently outperform the alternatives, with RoBERTa achieving 91-99\% F1-scores and accuracies across all classes.  
Notably, attention-augmented LSTMs with BERT embeddings approach transformer performance (up to 97\% F1-score) while training 2-3.5 times faster, whereas LSTMs using static embeddings fail to learn useful signals. These findings represent the first comprehensive benchmark for multi-class mental health detection, offering practical guidance on model selection and highlighting an accuracy–efficiency trade-off for real-world deployment of mental health NLP systems.

\end{abstract}


\section{Introduction}

Mental health conditions such as depression, anxiety, bipolar disorder, and complex post-traumatic stress disorder (CPTSD) are major public health concerns and often remain undiagnosed for years before they start severely affecting people's lives. The World Health Organization estimates that in 2019, nearly 970 million people worldwide lived with mental disorders, with anxiety and depression as the most common~\cite{WHO}. Meanwhile, millions of individuals have turned to social media platforms such as Reddit to share their struggles with mental illness in real time. Posts by these people offer a wealth of user-sourced words defining individuals' struggles, symptoms, emotional states, and affective experiences. If analyzed carefully, this kind of data could help identify mental health issues early and enable early, low-cost interventions with enormous potential to supplement existing healthcare systems~\cite{calvo2017natural}. This potential has spurred growing research on natural language processing (NLP) and mental health.
%
%

Early research largely focused on simple classification tasks, such as determining if a user is depressed, using traditional models or basic keyword features~\cite{coppersmith2014quantifying}. Those shallow approaches used conventional models such as Logistic Regressions, Support Vector Machines (SVM), topic modeling, and convolutional and recurrent neural networks (CNNs and RNNs) based on static word embeddings (e.g., Word2Vec and GloVe) to learn useful patterns~\cite{orabi2018deep, milne2016clpsych}. More recent studies have shifted toward more nuanced approaches to identify diverse mental health conditions across varied online expressions~\cite{milne2016clpsych}. While these models improved over rule-based systems, they were challenged by contextual ambiguity and long-distance dependencies. 
Recent studies~\cite{hasan2025generalmental} have shown that static embeddings often fail at handling the fine-grained, context-dependent language typically exhibited in mental health dialogue.

More advanced research in NLP, such as developing transformer-based models Bidirectional Encoder Representations from Transformers (BERT)~\cite{devlin2019bert} and its variants, promises more. They produce context-dependent embeddings, disambiguating word meaning in terms of consideration of local text, an important skill when dealing with emotionally charged or metaphorical language. Several publications demonstrated fine-tuned transformers outperforming standard models at detecting depression and anxiety from social media~\cite{chatterjee2021depression}. However, because most of this research was centered on one-disorder detection, very little is understood about how different model architectures perform on multi-class mental disorder classification problems. Moreover, despite the rapid adoption of transformers across mainstream NLP tasks, there exists no systematic study that (i) applies advanced transformer architectures to mental multi-condition classification and, simultaneously, (ii) benchmarks those models against modern Long Short-Term Memory (LSTM) variants under identical experimental conditions.

We present a complete empirical comparison of transformer and LSTM-based models for multi-class mental disorder prediction using Reddit data to address this gap. Building on our previous experience with binary classification, we generalize to six mental disorders (i.e., ADHD, anxiety, bipolar disorder, depression, CPTSD, and schizophrenia) and a control group. Using a rigorous data curation approach by self-disclosure and subreddit filtering, we obtain a vast dataset with dense mental health narratives.

\textbf{In brief, our contributions are fourfold:}

\begin{enumerate}
    \item We construct and release a well-annotated dataset of Reddit posts over various mental health issues, enabling fine-grained multi-class classification beyond binary detection.

    \item We endorse the annotation standard of the multi-class mental conditions posts through linguistic inspection, like divergence heat map and human judgmental analysis.

    \item We comparatively evaluate five transformer models alongside a range of LSTM models, including variants enhanced with attention and pre-trained embeddings. The results show that transformer models, especially RoBERTa, repeatedly outperform other approaches in accuracy and F1-score.

    \item We examine the performance of hybrid models, such as LSTMs augmented with BERT embeddings and an attention layer, offering a practical trade-off between performance and efficiency in low-resource settings.
\end{enumerate}

\section{Related Work}

Mental illness detection through social media text has gained traction in recent years, driven by the growing number of people openly sharing personal experiences online. Early approaches relied primarily on conventional machine learning techniques, often supplemented with hand-engineered linguistic features. For instance, Coppersmith et al.~\cite{coppersmith-etal-2015-adhd} used lexical pattern mining to identify depression and PTSD markers in tweets. Furthermore, Pirina et al.~\cite{pirina-coltekin-2018-identifying} analyzed Reddit posts on depression, showing that classifier performance hinged on the similarity between training and testing data. Although these studies provided foundational insights, their reliance on rigid methodologies limited their ability to capture the nuanced and diverse ways individuals express mental health struggles.

Deep learning marked significant progress in this area. Architectures like CNNs and RNNs, including LSTM models, proved better suited to identifying nuanced semantic relationships between words. Trotzek et al.~\cite{trotzek2020utilizing}, for instance, developed a character-CNN augmented with linguistic metadata features, achieving an F1-score of 0.63 for depressive content detection—a 9-point improvement over their SVM baseline. Orabi et al.~\cite{orabi2018deep} likewise showed LSTMs outperforming traditional methods in classifying depressive tweets. Subsequent work has refined these approaches through attention mechanisms, allowing models to weight emotionally salient words more effectively~\cite{ji2022suicidal}. Such innovations proved particularly valuable in high-stakes applications like suicide risk assessment, where the ability to identify emotionally significant terms is critical.

The transformer architectures, particularly BERT and its derivatives, have advanced NLP-based mental health classification by providing deep context-aware representations through a self-attention mechanism. Ezerceli et al.~\cite{ezerceli2024mental} achieved significantly better results with BERT-based classifiers than using CNN and LSTM-based methods for mental illness identification on social media. Hasan et al.~\cite{hasan2025generalmental} conducted an extensive evaluation of contextual embeddings and static embeddings in identifying suicidal ideation, reporting that contextual embeddings achieved an over 94\% F1-score and outperformed their counterpart, which attained around 80\% F1-score. Zhang et al.~\cite{zhang2022natural} performed an extensive review and indicated that several studies show the effectiveness of using transformer models for mental health detection.
These works demonstrate that transformers are superior at detecting complex and nuanced emotional cues in user content.

Despite these advances, existing literature has mostly focused on identifying a single mental disorder alone. Real-world clinical practice, however, requires discrimination among several mental illnesses at once, a considerably more challenging but practically meaningful task given symptom co-occurrence and linguistic similarities. The CLPsych~\cite{milne2016clpsych} attempted multi-class mental health severity classification from Reddit posts, with several teams achieving performance above baseline levels. However, this task primarily targeted severity categorization rather than differentiating among distinct disorders. Multi-task and hierarchical modeling approaches have recently begun investigating how to address this complexity more effectively. For instance, Lee et al.~\cite{lee2024detecting} employed multi-task learning methods to differentiate bipolar disorder from comparable illnesses like major depression, borrowing universal linguistic features to enhance precision.

Based on these previous attempts, our paper rigorously compares five advanced transformer models (RoBERTa, BERT, DistilBERT, ALBERT, ELECTRA) with a range of LSTM variants (attentional and non-attentional, contextual and static embeddings) on six various mental illnesses-ADHD, Anxiety, Bipolar Disorder, Depression, CPTSD, and Schizophrenia- vs. a Control group. To our knowledge, this rigorous comparative analysis across several model families and mental health classes is novel work, which provides open-ended implications for effective NLP-based mental health detection approaches and informs future clinical and computational research.

\section{Problem Definition}

The primary goal of this study is to classify social media posts collected from Reddit into specific mental health conditions using advanced deep learning models. We categorize posts into seven classes: ADHD, Anxiety, Bipolar Disorder, CPTSD, Depression, Schizophrenia, and a Control category (i.e., representing posts unrelated to any mental health condition).

\textbf{Problem Statement:} 
\textit{
Given an annotated dataset $D = \{d_1, d_2,\dots, d_n\}$ consisting of $n$ posts, we seek to train text classification models capable of assigning each post $d_i$ to one of seven predefined labels $d_i \in \{ADHD, Anxiety, Bipolar, CPTSD, Depression,\\ Schizophrenia, Control\}$.
}

To systematically address this task and provide clear insights on leveraging NLP techniques in automated mental health evaluation, we pose three central research questions:

\begin{enumerate}
    \item How well do transformer-based architectures perform in differentiating among multiple mental health conditions?

    \item To what extent do different embedding strategies, such as contextual embeddings (BERT) vs. static embeddings (GloVe, Word2Vec), impact the effectiveness of LSTM-based models, and how much does incorporating an attention mechanism contribute to their classification accuracy?

    \item Considering practical deployment, how robust and computationally efficient are transformer and LSTM models when trained on large-scale multi-class datasets?
\end{enumerate}

\section{Methodology}

\begin{figure}
    \centering
    
    \includegraphics[width=0.9\linewidth]{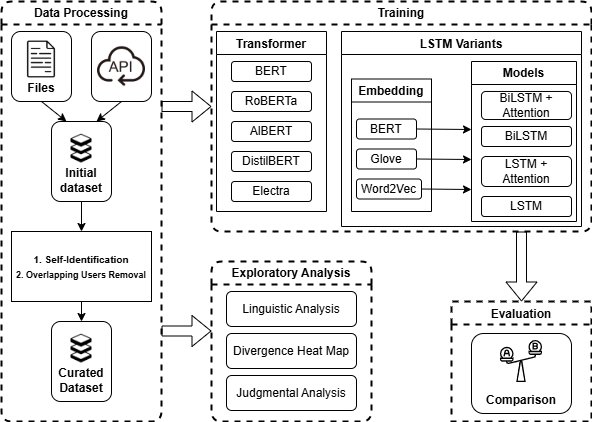}
    
    \caption{An Abstract of our Research Formation}
    \label{fig:research_formulation}
    \vspace{-4mm}
\end{figure}

Figure~\ref{fig:research_formulation} illustrates a broad summary of our research formulation, which begins with data processing and ends with model comparison.

\subsection{Dataset Creation}

\subsubsection{Data Collection \& Annotation}

We curated a new dataset of Reddit posts, sampled from six mental health-focused subreddits associated with specific diagnoses and several control subreddits unrelated to mental illness. The mental health forums are: r/ADHD, r/Anxiety, r/Bipolar, r/Depression, r/CPTSD, and r/Schizophrenia. Each of these subreddits centers on a particular mental disorder, where users share personal experiences and challenges related to their conditions. Posts from these subreddits served as candidate instances of the respective disorders, and accordingly, we considered labeling those posts with the corresponding mental health condition.

For the control group (users without diagnosed mental health conditions), we combined posts from general-interest subreddits unlikely to focus on mental health topics. These included r/Confidence, r/Politics, r/BicycleTouring, and r/Sports, covering themes such as self-improvement, news and politics, outdoor activities, and sports fandom—subjects with no inherent mental health bias. By diversifying the control group across multiple domains, we minimized topic-related biases, ensuring that classification relied more on linguistic markers of mental health rather than topic differences.


We retrieved posts using the Pushshift Reddit API\footnote{https://pushshift.io/}, which provides historical Reddit data. Our corpus contained all posts from the selected subreddits up to December 2022. This provided an initial corpus of over 3.3 million posts across six mental health and one control categories. Table~\ref{tab:mental_conditions_dataset} (Initial Dataset Statistics) includes the preliminary data volumes for each mental health subreddit. Further, we analyzed a divergence heat map and statistical annotation judgment using the posts of the labeled dataset to verify our annotation method as discussed in subsections \ref{subsubsec:linguistic_review} and \ref{subsubsec:human_judgmental_analysis}.

\begin{table}[!htb]
    \vspace{-2mm}
    \centering
    \caption{Dataset Statistics: Initial and Processed Counts of Posts for Mental Health Conditions}
    \label{tab:mental_conditions_dataset}
    \begin{tabular}{l|r|r}
        \hline
        \textbf{Subreddit} & \textbf{Initial (K)} & \textbf{Processed (K)} \\
        \hline
        ADHD & 643K & 220K \\
        Anxiety & 571K & 100K \\
        Bipolar & 303K & 85K \\
        CPTSD & 203K & 25K \\
        Depression & 1,516K & 191K \\
        Schizophrenia & 96K & 22K \\
        \hline
        \textbf{Total} & \textbf{3,332K} & \textbf{643K} \\
        \hline
    \end{tabular}
    \vspace{-2mm}
\end{table}

\subsubsection{Data Processing}

One of the key steps in our processing is to confirm that posts tagged with a certain condition are actually from users likely to have been affected by it, and not posts written by another user or generalized queries. For validation, we employed self-identification methods drawn from prior research~\cite{coppersmith-etal-2015-adhd, mitchell-etal-2015-quantifying}. Specifically, we designed regular expressions for the identification of overt self-diagnosis statements. We then matched these confirmed self-identified users with submissions in corresponding mental health subreddits.

Furthermore, we excluded users active in multiple mental health subreddits to maintain clear class boundaries, as their mixed contributions could introduce noise. Following Cohan et al.~\cite{cohan-etal-2018-smhd}, we removed all posts from users who participated in two or more mental health communities, ensuring each user contributed to only one diagnostic category. This approach yielded a well-separated dataset with minimal overlap. After applying self-identification and overlapping user removal techniques, we sampled 105,000 posts stratified evenly across subreddits to balance representation. This sample served as our primary dataset for model training and testing. 

We retained several snapshots of the data because different downstream tasks place varying demands on the text. For linguistic analyses, like part-of-speech (POS) tagging and divergence heat-map generation, we used a tokenized-only version that preserved the original word forms required by taggers. In contrast, features that depend on raw surface cues, such as character or token length and the extraction of multi-word expressions for model input, operated on the unaltered text to maintain full contextual details. This staged approach balances cleaning with fidelity, providing the most appropriate representation for each analytical objective.

\subsection{Exploratory Analysis}
Our exploratory procedure has two phases: (1) We examine posts categorically to find distinctive stylistic and lexical preferences that signal how the classes differ, and (2) We conduct a judgment-based audit, an inter-annotator agreement, to verify that the utilized labels are uniform and dependable before model building. Together, these stages provide qualitative data and quality assurance for the dataset.

\subsubsection{Linguistic Review}
\label{subsubsec:linguistic_review}

\begin{table}[htbp]
    \vspace{-2mm}
    \caption{Key Stylistic and POS Differences Across Classes}
    \label{tab:core-style-pos}
    \centering
    \small
    \begin{tabular}{lrrrrr}
        \hline
        \textbf{Class} &
        \begin{tabular}[c]{@{}c@{}}\textbf{URLs}\\[-2pt](avg)\end{tabular} &
        \begin{tabular}[c]{@{}c@{}}\textbf{Chars}\\[-2pt](avg)\end{tabular} &
        \begin{tabular}[c]{@{}c@{}}\textbf{Tokens}\\[-2pt](avg)\end{tabular} &
        \begin{tabular}[c]{@{}c@{}}\textbf{Nouns}\\[-2pt](avg)\end{tabular} &
        \begin{tabular}[c]{@{}c@{}}\textbf{Verbs}\\[-2pt](avg)\end{tabular} \\ 
        \hline
        Control        & 0.14 & 794.4 & 79.6 & 39.5 & 25.9 \\
        ADHD           & 0.04 & 1178.4 & 106.3 & 45.1 & 49.9 \\
        Anxiety        & 0.01 & 1243.5 & 111.8 & 45.6 & 54.1 \\
        Bipolar        & 0.02 & 967.9 & 86.9 & 35.6 & 41.7 \\
        CPTSD          & 0.04 & 1840.2 & 160.3 & 65.8 & 78.2 \\
        Depression     & 0.01 & 1523.97 & 133.5 & 52.6 & 68.9 \\
        Schizophrenia  & 0.07 & 1062.1 & 93.6 & 38.9 & 44.9 \\
        \hline
    \end{tabular}
\end{table}

The written posts exhibit distinct characteristics across classes, as shown in Table \ref{tab:core-style-pos}. Control posts are the shortest, averaging approximately 80 tokens and 800 characters, whereas CPTSD posts are nearly twice as long, with about 160 tokens and 1.8k characters. Verb usage also varies significantly among the groups. Control posts contain an average of 26 verbs, while ADHD and Anxiety posts include around 50. CPTSD posts lead with an average verb count of 78, suggesting more detailed, narrative descriptions. Noun usage similarly shows an upward trend across several mental health conditions.
Additionally, control users include external links more frequently (on avg. $0.14$ URLs per post), whereas all disorder groups display substantially lower linking behavior ($\leq0.07$ URLs per post). This reduced linking may suggest that individuals with mental health conditions tend to focus more on personal struggles rather than referencing external resources. 
These differences in post length, POS mix, and hyperlink usage indicate that labeled subjects exhibit different communication styles. However, posts related to mental health conditions also display nuanced differences among themselves.

\begin{figure}[htbp]
    \vspace{-1mm}
    \centering
    \includegraphics[width=0.9\linewidth]{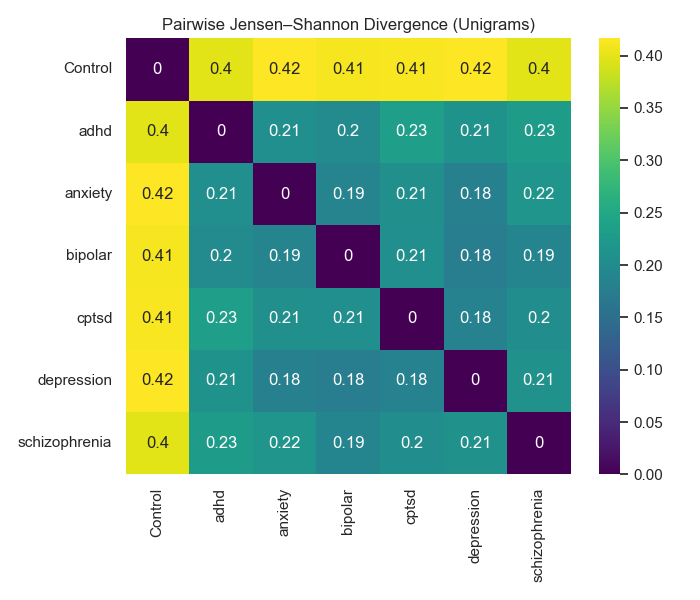}
    \caption{Pair-wise divergence map of each class}
    \label{fig:divergence_heat_map}
\end{figure}

Figure~\ref{fig:divergence_heat_map} depicts the pair-wise Jensen-Shannon (JS) divergence across the unigram language models of each class; values range from 0 (identical) to 1 (maximally different). As expected, posts in the Control group are lexically furthest from every disorder class (average $JS \approx 0.41$), whereas distances among disorders are lower ($JS \approx 0.18 - 0.23$). The smallest gap is observed between Depression and Bipolar ($JS = 0.18$), consistent with shared affective vocabulary, while ADHD and Anxiety retain a moderate divergence (0.21). The clear diagonal of zeros and the pronounced Control vs. disorder band confirm that each label captures a distinct lexical signal rather than random noise. Each class exhibits a distinctive linguistic footprint, justifying the quality of our annotations and downstream multi-class modeling.

\subsubsection{Human Judgmental Analysis}
\label{subsubsec:human_judgmental_analysis}

Human judgmental process strengthens our annotated dataset on top of the linguistic review described in section~\ref{subsubsec:linguistic_review}, and is a popular practice among researchers to validate annotation quality~\cite{Landis1977, hasan2021survey}. To verify the label consistency of our annotated dataset, we drew a stratified sample of 1,050 (1\%) posts and manually labeled them by two authors independently. Inter-annotator agreement was quantified with four standard statistics: percent agreement, Cohen's $\kappa$~\cite {Cohen1960}, Scott's $\pi$~\cite{scott1955}, and Krippendorff's $\alpha$~\cite{krippendorff2004}. Posts were judged against a concise rubric:

\begin{itemize}
  \item \textbf{Disorder labels (ADHD, Anxiety, Bipolar, CPTSD, Depression, Schizophrenia).} A post must contain a first-person disclosure of diagnosis or symptoms (e.g., ``\emph{I was diagnosed with bipolar}''), sustained descriptions of distress (panic attacks, manic episodes, persistent sadness, past traumatic experience, delusions, hallucination), usage of drugs commonly used to treat a certain condition, or explicit help-seeking related to that disorder.  
  
  \item \textbf{Control.} No self-referenced mental-health struggle; content is general (politics, sport, travel, hobbies) or problem-oriented without marked emotional strain.
\end{itemize}

\begin{table}[htbp]
    \vspace{-4mm}
    \caption{Inter-Annotator Agreement Scores}
    \label{tab:inter_annotator_agreement_score}
    \centering
    \begin{tabular}{l|cc}
    \hline
    \textbf{Metric} & \textbf{Annotator 1} & \textbf{Annotator 2} \\
    \hline
    Percent agreement & 93.33\% & 92.19\% \\
    Cohen's~$\kappa$  & 0.922 & 0.909 \\
    Scott's~$\pi$     & 0.922 & 0.909 \\
    Krippendorff's~$\alpha$ & 0.922 & 0.909 \\
    \hline
    \end{tabular}
    \vspace{-2mm}
\end{table}

Table~\ref{tab:inter_annotator_agreement_score} reports the agreement scores among human annotators. All inter-annotator agreement scores for both annotators exceed the ``almost-perfect'' threshold of 0.80 recommended by Landis et al. \cite{Landis1977}. These figures attest that the annotation guidelines are sufficiently clear for independent raters and that the labels can be trusted for downstream modeling.

\section{Classification Models}

In this work, we explore two broad categories of deep learning models to classify posts under several mental health conditions: transformer-based and LSTM-based models. These two categories employ various architectures and embedding strategies, explained below.

\subsection{Transformer-based Models}

We adapted various cutting-edge transformer models to the multi-class mental health detection task. Our choice consists of five popular models to address a range of model complexities and training strategies: 
 \textbf{(1)} BERT-base (uncased) is a 12-layer bidirectional transformer pretrained on large English data sets, known for its contextualized word representation and robustness to seize the subtleties of text semantics~\cite{devlin2019bert}.
 \textbf{(2)} RoBERTa is a fine-tuned, maximally optimized version of BERT with far greater quantities of data and superior training practices, which achieves state-of-the-art performance on various NLP tasks.
 \textbf{(3)} DistilBERT offers a light, distilled version of BERT with fewer parameters and layers but on-par performance, with much less computational resources.
 \textbf{(4)} ALBERT introduces parameter-reduction methods in factorized embedding parametrization form and layer sharing, enabling fine-tuning under minimal memory~\cite{Lan2019ALBERTAL}.
 \textbf{(5)} ELECTRA employs novel pretraining in the discriminator-generator framework. Rather than predicting the masked words (as in BERT), it identifies replaced tokens, which enhances pretraining performance and downstream accuracy~\cite{Clark2020ELECTRA}.

\subsection{LSTM-based Models}

The second category consists of LSTM architectures of several configurations. First, unidirectional LSTM consumes the text sequentially from start to finish, embedding contextual information as the sequence proceeds. Second, bidirectional LSTM (BiLSTM) encodes in both forward and reverse directions simultaneously so that the model can consider the entire context in which each token is situated and potentially acquire richer semantics. 

We further augmented these LSTM models with an optional attention mechanism. Instead of compressing the sequence based on the final hidden state only, the attention layer computes learnable weights for every hidden state along the sequence, creating a weighted sum that gives more weight to important regions in the text. The mechanism can drastically improve performance by allowing the model to focus on important phrases indicative of particular mental states.

In all LSTM variants, we experimented with three embedding methods to represent textual inputs:
(1) \textbf{Contextual Embeddings (BERT):} Using BERT as a feature extractor, we obtained token-level contextual embeddings from a pre-trained BERT-base (uncased) model without additional fine-tuning. This method combines transformer-based embedding's high semantic awareness and the sequence modeling capability of LSTM networks.
(2) \textbf{Static Embeddings (GloVe):} We used 300-dimensional pre-trained GloVe embeddings trained on Common Crawl (840b tokens), providing fixed vector representations of each word regardless of the context of occurrence.
(3) \textbf{Static Embeddings (Word2Vec):} We also explored the utilization of 300-dimensional Word2Vec embeddings pre-trained on the Google News corpus, providing an alternative static representation using an alternative training corpus and vocabulary.

For efficiency and regularization, the hidden layer size in LSTM networks was set to 128 units per direction (256 for BiLSTM), and dropout was fixed at 0.2. In attention models, the attention layer was a plain feed-forward neural network that produced a scalar weight for each hidden state of the LSTM sequence.


\subsection{Computational Cost Analysis}

Table~\ref{tab:computational_complexity} highlights the significant disparity in the computational cost and memory usage in the transformer-based models compared to the leaner LSTM-based models. As expected, the transformer models consume at least 89.5 billion floating-point operations (FLOPs) in a single forward pass. LSTMs, however, have below 0.01 billion FLOPS and use much less GPU memory. Hence, LSTM-based pipelines provide a far more resource-friendly alternative that can be more viable in real-time or environment-constrained environments.

\begin{table}[htbp]
    \vspace{-2mm}
    \centering
    \resizebox{\columnwidth}{!}{%
    \begin{threeparttable}
        \caption{Comparison of FLOPs, Parameters, and Peak GPU Memory Usage Across Transformer-based and LSTM-based Models}
        \label{tab:computational_complexity}
        \begin{tabular}{|l|c|c|c|}
        \hline
        \textbf{Model} & \textbf{FLOPs} & \textbf{Parameters} & \textbf{Peak GPU} \\
         & \textbf{(Billions)} & \textbf{(Millions)} & \textbf{Memory (GB)} \\
        \hline
        Full BERT & 179.0 & 109.5 & 1.72 \\
        Full RoBERTa & 179.0 & 124.6 & 1.76 \\
        Full DistilBERT & 89.5 & 67.0 & 0.91 \\
        Full ELECTRA & 179.0 & 109.5 & 1.72 \\
        Full ALBERT & 179.2 & 11.7 & 2.18 \\
        \hline
        BERT-Embedding+BiLSTM+Attn & 0.0095 & 24.8 & 0.18 \\
        BERT-Embedding+BiLSTM & 0.0089 & 24.8 & 0.18 \\
        BERT-Embedding+LSTM+Attn & 0.0087 & 24.3 & 0.15 \\
        BERT-Embedding+LSTM & 0.0084 & 24.3 & 0.15 \\
        \hline
        GloVe+BiLSTM+Attn & 0.0016 & 0.51 & 0.07 \\
        GloVe+BiLSTM & 0.0011 & 0.51 & 0.07 \\
        GloVe+LSTM+Attn & 0.0008 & 0.26 & 0.05 \\
        GloVe+LSTM & 0.00055 & 0.26 & 0.05 \\
        \hline
        Word2Vec+BiLSTM+Attn & 0.0016 & 0.51 & 0.07 \\
        Word2Vec+BiLSTM & 0.0011 & 0.51 & 0.07 \\
        Word2Vec+LSTM+Attn & 0.0008 & 0.26 & 0.05 \\
        Word2Vec+LSTM & 0.00055 & 0.26 & 0.05 \\
        \hline
        \end{tabular}

        \begin{tablenotes}[flushleft]
            \centering
                \item[*] Experimented with batch 16 and sequence length 128
        \end{tablenotes}
    \end{threeparttable}
    }
    \vspace{-4mm}
\end{table}

\section{Experimental Setup}

We evaluated our models using 5-fold stratified cross-validation to generate good and reliable performance estimates. The dataset was first shuffled randomly and stratified into an 80/20 ratio while maintaining class distributions. We kept a 20\% subset as a separate hold-out test set that was not used at training time. The remaining 80\% was split into five stratified folds. All model configurations were cross-validated and trained five times, each with a distinct fold for validation and the other four being utilized for training, resulting in five trained versions of each model. Final predictions on the hold-out test set were made by combining predictions from each of the five trained models through majority voting, resolving ties at random. This ensemble approach greatly improved prediction stability and reduced biases inherent in individual models.

Parameters were adjusted at training time with the AdamW optimizer (Adam with weight decay)~\cite{loshchilov2018decoupled}. Hyperparameters were optimized automatically with Ray Tune\footnote{https://www.ray.io/}, and best performance was universally achieved with a small learning rate $10^{-6}$ and weight decay $10^{-2}$. Early stopping with a patience of 5 epochs to prevent overfitting and improve generalization was also used, which terminated training when validation loss stopped improving. Additionally, we employed standard performance metrics to assess model performance: accuracy, precision, recall, and F1-score. For this multi-class scenario, we primarily considered macro-averaged values with caution to balance the evaluation of all classes. We performed class-level analysis to differentiate model performance across distinct mental health categories.

\begin{table*}[ht]
    \centering
    \begin{threeparttable}
        \caption{An Overview of Models' Performance Based on Accuracy, F1 Score, Precision, and Recall
        }
        \label{tab:experimental_results}

        \begin{scriptsize}
        \begin{tabular}{|c|c||c|c|c|c|c|c|c|c||c|}
        \hline
        \textbf{Embedding} & \textbf{Model} & \textbf{Metric} & \textbf{Control} & \textbf{ADHD} & \textbf{Anxiety} & \textbf{Bipolar} & \textbf{CPTSD} & \textbf{Depression} & \textbf{Schizophrenia} & \textbf{Time (hrs)\tnote{*}} \\ \hline
        
        \multirow{4}{*}{RoBERTa} & \multirow{4}{*}{RoBERTa}
         & Accuracy & 98.87 & 97.67 & 94.00 & 92.17 & 95.57 & 92.33 & 95.67 & \multirow{4}{*}{3.99} \\ 
         & & F1-score & 99.20 & 97.78 & 94.03 & 92.68 & 95.74 & 91.46 & 95.41 &  \\ 
         & & Precision & 99.53 & 97.90 & 94.06 & 93.19 & 95.92 & 90.61 & 95.16 &  \\ 
         & & Recall & 98.87 & 97.67 & 94.00 & 92.17 & 95.57 & 92.33 & 95.67 &  \\ 
         \hline
        
        \multirow{4}{*}{DistilBERT} & \multirow{4}{*}{DistilBERT} 
         & Accuracy & 98.67 & 97.93 & 94.03 & 90.13 & 95.53 & 91.67 & 95.83 & \multirow{4}{*}{2.37} \\ 
         & & F1-score & 98.91 & 97.53 & 93.64 & 92.07 & 95.33 & 90.94 & 95.36 &  \\ 
         & & Precision & 99.16 & 97.12 & 93.26 & 94.08 & 95.12 & 90.22 & 94.88 &  \\ 
         & & Recall & 98.67 & 97.93 & 94.03 & 90.13 & 95.53 & 91.67 & 95.83 &  \\ 
         \hline
        
        \multirow{4}{*}{ALBERT} & \multirow{4}{*}{ALBERT} 
         & Accuracy & 98.53 & 97.80 & 93.23 & 90.83 & 95.03 & 90.17 & 95.63 & \multirow{4}{*}{3.19} \\
         & & F1-score & 98.65 & 97.46 & 93.02 & 91.91 & 95.02 & 90.21 & 94.94 &  \\
         & & Precision & 98.76 & 97.12 & 92.80 & 93.00 & 95.00 & 90.26 & 94.25 &  \\
         & & Recall & 98.53 & 97.80 & 93.23 & 90.83 & 95.03 & 90.17 & 95.63 &  \\
         \hline
        
        \multirow{4}{*}{BERT} & \multirow{4}{*}{BERT} 
         & Accuracy & 98.50 & 97.43 & 94.07 & 91.70 & 95.57 & 91.63 & 95.27 & \multirow{4}{*}{4.10} \\
         & & F1-score & 98.85 & 97.38 & 93.52 & 92.18 & 95.82 & 91.10 & 95.35 &  \\
         & & Precision & 99.19 & 97.34 & 92.98 & 92.66 & 96.08 & 90.58 & 95.43 &  \\
         & & Recall & 98.50 & 97.43 & 94.07 & 91.70 & 95.57 & 91.63 & 95.27 &  \\
         \hline
        
        \multirow{4}{*}{ELECTRA} & \multirow{4}{*}{ELECTRA} 
         & Accuracy & 98.63 & 97.00 & 94.03 & 90.47 & 94.70 & 93.70 & 96.47 & \multirow{4}{*}{4.48} \\
         & & F1-score & 99.01 & 97.45 & 93.78 & 92.30 & 95.88 & 90.91 & 95.76 &  \\
         & & Precision & 99.40 & 97.91 & 93.53 & 94.20 & 97.10 & 88.29 & 95.07 &  \\
         & & Recall & 98.63 & 97.00 & 94.03 & 90.47 & 94.70 & 93.70 & 96.47 &  \\
         \hline 
        
        \multirow{4}{*}{BERT} & \multirow{4}{*}{BiLSTM+Attn} 
         & Accuracy & 98.17 & 96.20 & 91.27 & 89.73 & 94.00 & 89.23 & 94.40 & \multirow{4}{*}{1.04} \\
         & & F1-score & 97.84 & 96.28 & 91.54 & 90.75 & 94.36 & 88.03 & 94.23 &  \\
         & & Precision & 97.52 & 96.36 & 91.82 & 91.78 & 94.73 & 86.86 & 94.06 &  \\
         & & Recall & 98.17 & 96.20 & 91.27 & 89.73 & 94.00 & 89.23 & 94.40 &  \\
         \hline 
        
        \multirow{4}{*}{BERT} & \multirow{4}{*}{BiLSTM} 
         & Accuracy & 90.27 & 28.47 & 33.33 & 21.53 & 74.70 & 22.33 & 25.50 & \multirow{4}{*}{0.95} \\
         & & F1-score & 67.38 & 40.04 & 25.32 & 29.15 & 55.57 & 26.94 & 35.37 &  \\
         & & Precision & 53.75 & 67.46 & 20.42 & 45.11 & 44.24 & 33.94 & 57.69 &  \\
         & & Recall & 90.27 & 28.47 & 33.33 & 21.53 & 74.70 & 22.33 & 25.50 &  \\
         \hline
        
        \multirow{4}{*}{BERT} & \multirow{4}{*}{LSTM+Attn} 
         & Accuracy & 98.00 & 96.27 & 90.57 & 87.13 & 92.53 & 88.67 & 92.63 & \multirow{4}{*}{1.21} \\
         & & F1-score & 97.85 & 96.30 & 90.13 & 88.90 & 93.64 & 85.99 & 93.15 &  \\
         & & Precision & 97.71 & 96.33 & 89.70 & 90.73 & 94.78 & 83.46 & 93.66 &  \\
         & & Recall & 98.00 & 96.27 & 90.57 & 87.13 & 92.53 & 88.67 & 92.63 &  \\
         \hline
        
         \multirow{4}{*}{BERT} & \multirow{4}{*}{LSTM} 
         & Accuracy & 94.10 & 59.57 & 62.83 & 9.93 & 54.47 & 9.27 & 4.87 & \multirow{4}{*}{1.49} \\
         & & F1-score & 62.37 & 55.70 & 38.69 & 17.14 & 51.54 & 14.76 & 9.12 &  \\
         & & Precision & 46.65 & 52.30 & 27.95 & 62.47 & 48.91 & 36.29 & 71.92 &  \\
         & & Recall & 94.10 & 59.57 & 62.83 & 9.93 & 54.47 & 9.27 & 4.87 &  \\
        \hline

        \end{tabular}
        \end{scriptsize}

        \begin{tablenotes}[flushleft]
            \centering
                \item[*] Approx. training time to achieve a best-performing model; this may vary depending on hardware specifications.
        \end{tablenotes}
    \end{threeparttable}
    \vspace{-4mm}
\end{table*}

All experiments were run using PyTorch and HuggingFace Transformers on a computational setup consisting of two NVIDIA A100 GPUs (Ampere architecture, 40 GB memory per GPU) and 512 GB system RAM. This robust infrastructure facilitated parallel fine-tuning of large transformer models. Training times were also measured to provide actual insights into the computational needs of each model architecture.

\section{Discussion}

After training all models under the 5-fold cross-validation scheme, we aggregated the results to compare their performance on the hold-out test set. Table~\ref{tab:experimental_results} summarizes the performance in the key metrics (i.e., accuracy, macro-averaged precision, recall, F1, and training time) for each model and embedding configuration except the static embeddings (i.e., Word2Vec, GloVe). Here, the training time metric is the average time required to achieve a best-performing model in the early stopping setup, calculated by taking the mean training time of 5-fold models. We also experimented with LSTM-based models paired with static embeddings. However, these approaches had extremely low predictive capabilities, always producing 0\% or poor F1-scores (less than 30\%) for every class. This vividly indicates the shortcomings of applying static word embeddings to identify the subtle and context-dependent linguistic patterns in mental health conversations. Hence, our sophisticated comparative analysis focuses on transformer-based and contextually enriched LSTM models due to their much improved performance.

\subsection{How well do transformer-based architectures perform in differentiating among multiple mental health conditions?}

Transformer models exhibited an impressive capacity to differentiate among mental health conditions, as easily testified by their steady, statistically significant accuracy across our dataset. RoBERTa, for example, was highly accurate with accuracies of 92.17\% (Bipolar) to 98.87\% (Control) testified to by equally high F1-scores of 91.46\% to 99.20\%. Other transformer models such as BERT, DistilBERT, ALBERT, and ELECTRA also demonstrated consistent high accuracies (90\%–98\%) and high F1-scores akin to RoBERTa.

This impressive capability stems from the inbuilt strength of transformer models: their attention mechanism allows the models to distinguish very subtle and crucial contextual cues of users' postings, much needed for discrimination among similar mental states like Anxiety vs. Depression or Bipolar vs. CPTSD. In addition, the extensive pre-training on multilingual datasets allows transformer variants to achieve dense linguistic knowledge that helps them to identify extremely intricate and frequently vague emotional language common in mental health conversations.

Furthermore, transformers have a functional advantage in terms of precision. For instance, DistilBERT, with its smaller model size, achieved similar performance, slightly less than RoBERTa's, yet substantially lower computational cost in terms of overhead, demonstrating that functional efficiency can be sacrificed for high precision.

Overall, transformer models demonstrate dual advantages for multi-class mental health classification: (1) superior accuracy and nuanced discrimination between conditions, and (2) adaptable control over model complexity and computational requirements. These characteristics position transformer architectures as suitable for developing scalable, deployment-ready social media-based mental health detection systems.

\subsection{To what extent do different embedding strategies, such as contextual embeddings (BERT) vs. static embeddings (GloVe, Word2Vec), impact the effectiveness of LSTM-based models, and how much does incorporating an attention mechanism contribute to their classification accuracy?}

Our experimental results show that word embeddings significantly impact the performance of the LSTM-based models, with contextual (BERT) vs. static embeddings (GloVe, Word2Vec) being a distinguishing factor.

BERT contextual embeddings consistently improved classification performance. For example, the attentional BiLSTM with BERT embeddings achieved accuracy from 89.23\% (Depression) to 98.17\% (Control), with remarkable F1-scores (88.03\%–97.84\%). Performance robustness owes a great deal to the ability of BERT embeddings to manipulate the meanings of words in terms of context so that LSTM models can understand sensitive emotional nuances critical to identifying specific mental health illnesses.

On the other hand, static embedding-based LSTM models like GloVe and Word2Vec performed terribly. More specifically, they would over-emphasize some classes at the cost of drastic performance degradation. For instance, F1-score and accuracy were 0\% for almost every category, meaning static embeddings could not provide the level of context needed to distinguish subtle emotional expressions, thus severely weakening their abilities.

Adding an attention layer in the end to BERT-embedded LSTM variants improved model accuracy even more. For instance, adding attention to a BERT-extended BiLSTM improved accuracy much more than its non-attentional version, at times by as much as 60\% or more, particularly in difficult categories like Bipolar and Schizophrenia. The attention layer allowed the model to dynamically focus on important sentences or symptoms of each post, greatly improving its ability to extract important signals in long and intricate stories.

In short, our findings strictly confirm that context embedding utilization significantly improves the performance of LSTMs, and attention mechanisms further enhance their accuracy by governing the model to focus on the most relevant features of the text. The hybrid model thus yields an efficient solution for multi-class mental illness classification from social media posts.

\subsection{Considering practical deployment, how robust and computationally efficient are transformer and LSTM models when trained on large-scale multi-class datasets?}

From a practical deployment perspective, our experiments determine clear trade-offs between transformer-based and LSTM-based models for robustness, computational efficiency, and accuracy in classifying mental conditions. Transformer models, i.e., RoBERTa and BERT, were the strongest and achieved the highest accuracy, with all classes consistently between 90\% and 99\%. Such high-performing transformers are computationally costly, though. For example, it took RoBERTa around 4 hours to train on our infrastructure, and BERT took around the same time. Even DistilBERT, a resource-efficient transformer variant with comparable accuracy (90\%–98\%), took around 2.4 hours, which is still considerable computational effort.

In comparison, LSTM models—particularly those extended with contextual BERT embeddings and attention mechanisms—provided similar performance with significantly lighter computational overhead. Such attention-enhanced LSTMs would typically take about 1 hour to train, approximately 2.5–4 times faster than transformer models. Though their accuracy (i.e., 89\%–98\%) was somewhat inferior to that of transformers, this accuracy compromise is acceptable considering the huge training efficiency advantage.

On the other hand, less complex LSTMs with static embeddings (Word2Vec and GloVe) failed to capture the subtleties of mental health language (e.g., F1-scores usually near 0\% for most of the experimental cases), resulting in worse classification performance. Their low performance highlights that simplifying model complexity severely undercuts predictive validity, an unacceptable trade-off for real-world applications.

Thus, practical deployment environments must carefully balance these considerations. Where extremely high accuracy is of prime concern and resources are no object, transformer models are preferred. In resource-limited environments or applications, where quicker model training and deployment are necessary, attention-augmented LSTM models with contextual embeddings offer an appealing, pragmatic trade-off with tolerable accuracy without prohibitive computational cost.

\section{Conclusion and Future Work}

Our systematic comparison of transformer and LSTM architectures for multi-class mental health classification from Reddit data yielded three principal findings:
(1) \textbf{Transformer Superiority}: Transformer models demonstrated consistent performance advantages, with RoBERTa achieving 92\%–99\% accuracy across all diagnostic categories. Comparable performance from BERT, DistilBERT, ALBERT, and ELECTRA confirms transformers' exceptional capability in detecting nuanced mental health language patterns;
(2) \textbf{LSTM Competitiveness}: Attention-enhanced LSTMs with BERT contextual embeddings approached transformer-level accuracy (89\%–98\%), whereas static embeddings (GloVe, Word2Vec) failed catastrophically, conclusively demonstrating the necessity of contextual representations for mental health text analysis;
(3) \textbf{Efficiency Trade-offs}: Transformers incurred three to four times greater computational costs than optimized LSTMs, establishing an important accuracy/efficiency trade-off for resource-constrained deployments.

Even with such encouraging outcomes, the findings suggest several areas of progress and advancement for future work. Firstly, we plan to investigate domain-specific pretraining techniques to facilitate model generalizability over multiple social media platforms. This encompasses pretraining transformer models over large-scale domain-specific datasets for mental health in a potential cross-platform manner. Moreover, future studies will explore integrating user-oriented multi-modal signals, such as user interaction patterns, to have better predictive accuracy and validity. Finally, we intend to use explainable AI techniques, including attention visualization and feature-importance analysis, which allow clinicians and moderators to comprehend model outputs better and gain confidence in the auto-system.

In conclusion, this study advances our understanding of efficient deep learning solutions to multi-class mental health classification. With continued innovation in these directions, we look forward to significantly contributing to developing beneficial, interpretable, and effective systems that enable early intervention and indirectly improve mental health outcomes worldwide.

\bibliographystyle{ieeetr}
\bibliography{references}

\end{document}